\documentclass{article}

\usepackage{arxiv}

\usepackage[utf8]{inputenc} 
\usepackage[T1]{fontenc}    
\usepackage{hyperref}       
\usepackage{url}            
\usepackage{booktabs}       
\usepackage{amsfonts}       
\usepackage{nicefrac}       
\usepackage{microtype}      
\usepackage{multicol}
\usepackage{graphicx}
\usepackage{float}
\usepackage{lipsum}
\usepackage{amsmath}
\usepackage{datetime}

\graphicspath{ {./images/} }

\title{Application of Deep Self-Attention in Knowledge Tracing}

\author{
 Junhao Zeng \\
  College of Computer Science and Technology\\
  Zhejiang University\\
  \texttt{fusion@zju.edu.cn} \\
   \And
 Qingchun Zhang \\
  College of Computer Science and Technology\\
  Zhejiang University\\
  \texttt{ekkoruse@zju.edu.cn} \\
  \And
 Ning Xie \\
  College of Computer Science and Technology\\
  Zhejiang University\\
  \texttt{tonyx@zju.edu.cn} \\
  \And
 Bochun Yang \\
  College of Computer Science and Technology\\
  Zhejiang University\\
  \texttt{Ybc@zju.edu.cn} \\
  
}

\begin{document}
\maketitle
\begin{abstract}
The development of intelligent tutoring system has greatly influenced the way students learn and practice, which increases their learning efficiency. The intelligent tutoring system must model learners' mastery of the knowledge before providing feedback and advices to learners, so one class of  algorithm called "knowledge tracing" is surely important. 
This paper proposed Deep Self-Attentive Knowledge Tracing (DSAKT) based on the data of PTA, an online assessment system used by students in many universities in China, to help these students learn more efficiently. 
Experimentation on the data of PTA shows that DSAKT outperforms the other models for knowledge tracing an improvement of AUC by 2.1\% on average, and this model also has a good performance on the ASSIST dataset

\end{abstract}

\keywords{Knowledge Tracing\and Deep learning\and Self-attention}
\vskip 0.6in

\begin{multicols}{2}
\section{INTRODUCTION}

Intelligent Tutoring System (ITS) is a type of computer system used for learning assistance, which can provide timely feedback to learners to help them learn more efficiently. To meet this position, ITS need to model learner's mastery of knowledge concepts and relations between different knowledge concepts. Therefore, one class of algorithm called "knowledge tracing" (KT) were invented and used in all kinds of models. 
During last tens of years, ITS has developed a lot. Nowadays they are playing more and more important roles in the field of online education. Thus, how to increase the accuracy of these models is an important problem. 

Given a student's response to a series of exercises $E_1,\dots,E_{n-1}$, the KT task can be described as to predict the probability that the student will be able to give the correct answer to the next exercise $E_n$. Here, the interactions between a student and a series of exercises are represented by a sequence I, in which each element is a tuple $I_i= (E_i,R_i)$, where $E_i$ denotes the exercise information of the $i$th exercise, and $R_i$ denotes the student’s response to exercise $E_i$. $r_i\in\{0,1\}$ is a part of $R_i$, which indicates whether the student is successful in finishing exercise $E_i$. Generally speaking, our model predicts the following probability: 
\begin{equation*}
    P(r_n=1|I_1,\dots,I_{n-1},E_n)
\end{equation*}

Programming Teaching Assistant (\href{http://pintia.cn}{PTA})\footnote{\url{http://pintia.cn}}is an teaching assistant platform used by students of many universities in China, on which students can practice programming freely. College teachers may also assign programming tasks to students on PTA platform to improve students' programming ability. Both above produces a large number of student practice data. These data can be used for knowledge tracing (KT) to model students' mastery situation of the knowledge concepts, then implement an ITS on PTA. 
Different from the general test platform, PTA allows students to submit the same exercise repeatedly until they get it right, which leads to a great increase in the interaction in PTA data. And college teachers generally use PTA to test students in the process of teaching, students' mastery of knowledge is gradually rising in the completion of exercises, and it causes that the interaction of a user on a specific exercise in PTA data showing the situation of "wrong before right"
These differences make it difficult for traditional models like DKT or SAKT to deal, but at the same time, represented the differences between programming problems and other problems. Therefore, these differences are valuable and necessary to deal with. Finally, we proposed a method based on the ideas of SAKT, which applied a more complete self-attention layer and achieved better results on the data of PTA.
\begin{figure}[H]
\centering
\includegraphics[width=0.49\textwidth]{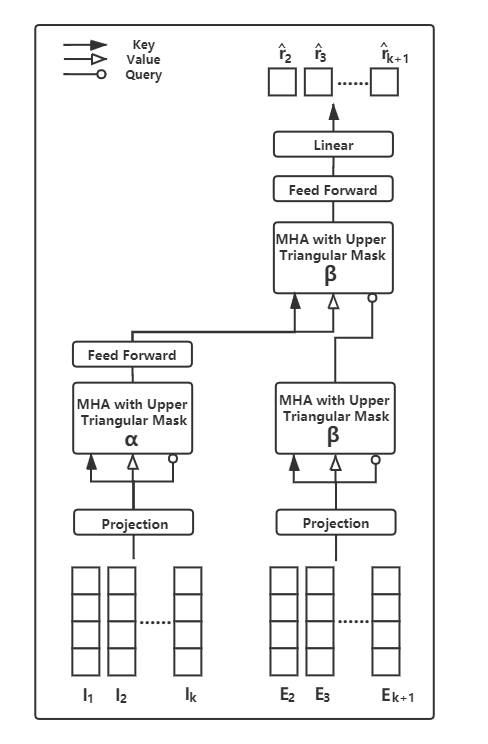}
\caption{The architecture of our proposed model, DSAKT}
\label{DSAKT_architecture}
\end{figure}

In this paper we improved SAKT and proposed an encoder-decoder based model named Deep Self-Attentive Knowledge Tracing (DSAKT). The main improvement is that we use the Multi-Head Attention (MHA) \cite{Attention} layer twice with the same weight in decoder, which make it possible to remain the relations between knowledge concepts completely until making prediction. Thus, the proposed DSAKT method gains the average AUC of PTA datasets is improved by 2.1\%. In addition, the self-attention based component is suitable for parallelism, which makes it one order of magnitude faster than the RNN based model.

\section{RELATED WORKS}
These years as neural network developing rapidly, traditional knowledge tracing algorithm like Bayesian Knowledge Tracing (BKT) \cite{BKT} and \cite{col1, col2} collaborative filtering has been studied extensively \cite{tradition1, tradition2, tradition3} over time and gradually been replaced by methods using neural network. 

Deep Knowledge Tracing (DKT) proposed in \cite{DKT} is one of the important algorithms, which is invented by researchers in Stanford University and developed rapidly in nearly several years. DKT and the variants \cite{DKT-RNN,FDKT}  creatively uses Recurrent Neural Network (RNN) to do modeling. DKT uses one-hot encoding to represent the results of practicing, which is simple to implement but may cause the input vector too long and sparse. Thus compress-sensing is used for solving this problem. However, in real life the number of knowledge concepts students meet is not huge, which causes the data too sparse, and in this case DKT cannot work well. 

Dynamic Key-value memory network (DKVMN) \cite{DKVMN}, although utilized Memory Augmented Neural Network \cite{MANN} for Knowledge Tracing to become more interpretable \cite{DKT-Howdeep}, still faces the same problem as DKT \cite{DKT-sparse}. 

DKT has become an important and mainstream knowledge tracing method, and is continuously being improved. Recently, based on the original DKT model and its variants, many researchers have made some valuable achievements \cite{FDKT,DKVMNR,VDKT}.

Self-Attentive Knowledge Tracing (SAKT) \cite{SAKT} aims to deal with the problem of sparsity of the real-world data. The authors proposed a self-attention \cite{Attention} based approach, which pays less attention on past activities and identifies the relevance between knowledge concepts when making predictions.

Due to the particularity of the data of PTA, we studied SAKT and made some creative improvement on its embedding layer and its transformer, implemented a new algorithm named Deep Self Attentive Knowledge Tracing. 

\section{PROPOSED MODEL}

\subsection{Input Representation}
DSAKT takes a series of interactions $I=[I_1,\dots{},I_k]$  with the length of k and the next exercise $E_k$ as input, and predicts probability of the student will make $E_{k+1}$ correctly. In order to facilitate subsequent data processing, we transform the interaction into:
\begin{equation*}
    I_i=r_i*e+E_i
\end{equation*}
Where $e$ denotes the response bias number, usually the total number of exercises. Therefore, the final input is:
\begin{equation*}
\left\{
\begin{aligned}
I&=[I_1,I_2,\dots,I_k] \\
{E}'&=[E_2,E_3,\dots,E_{k+1}] 
\end{aligned}
\right.
\end{equation*}

\subsection{Embedding layer}
The embedding layer in DSAKT is consist of three embedding matrices, $\hat{E}\in\mathbb{R}^{e\ast d}$, $\hat{P}\in\mathbb{R}^{k\ast d}$, $\hat{I}\in\mathbb{R}^{2e\ast d}$, where $e$ is the total number of exercise, $k$ is the number of the sequence length, and $d$ is the latent dimension.

\paragraph{Position Encoding}
The above $\hat{P}\in\mathbb{R}^{k\ast d}$ is a fixed and invariant coordinate embedding, which is obtained by calculating the sine position embedding vector defined in \cite{position}:
\begin{equation*}
\hat{P}_{i,j}=
    \left\{
    \begin{aligned}
    &\sin{(i/10000^{j/d})} &\mbox{j mod 2 = 0} \\
    &\cos{(i/10000^{(j-1)/d})} &\mbox{j mod 2 = 1}
    \end{aligned}
    \right.
\end{equation*}
Where $1\leq i\leq k$ is the location and $1\leq j\leq d$ is the dimension. In other words, each dimension of position embedding corresponds to a sine-curve.

$\hat{I}$ and $\hat{P}$ transform the obtained input sequence $I$ into a vector in latent space by embedding the interaction and position respectively, and then adding the embedding vectors together. In order to embed $E_{k+1}$ in to a vector in the latent space of the same shape, we use the information of last $k-1$ exercises in the interaction to splice with $E_{k+1}$ and transform them through $\hat{E}$ and $\hat{P}$.

\paragraph{Projection}
After embedding, we use two projection matrices $W^{I}\in\mathbb{R}^{d\ast d}$, $W^{E}\in\mathbb{R}^{d\ast d}$ to project the respective vectors to different space linearly. The output of embedding layer are embedded vectors $\tilde{I}\in\mathbb{R}^{k\ast d}$, $\tilde{E}\in\mathbb{R}^{k\ast d}$

\subsection{Deep Self-attentive Encoder-Decoder}
Our proposed model DSAKT, based on the multi-head self-attention architecture proposed in \cite{Attention}, consist of an improved encoder-decoder mechanism as shown in Figure \ref{DSAKT_architecture}. The encoder takes the embedded vector $\tilde{I}=[\tilde{I_1},\tilde{I_2},\dots,\tilde{I_k}]$ as input and feeds the processed output $\tilde{T}=[\tilde{T_1},\tilde{T_2},\dots,\tilde{T_k}]$ to the decoder. The decoder receives $\tilde{T}$ and another sequential input $\tilde{E}=[\tilde{E_1},\tilde{E_2},\dots,\tilde{E_k}]$ of the exercise embedding vector and generates the predicted response $\tilde{r}=[\tilde{r_1},\tilde{r_2},\dots,\tilde{r_k}]$.
\begin{equation*}
\left\{
\begin{aligned}
\tilde{T}&={\rm Encoder}(\tilde{I})\\
\tilde{r}&={\rm Decoder}(\tilde{E}, \tilde{T})
\end{aligned}
\right.
\end{equation*}

\paragraph{Multi-head Attention}
The most fundamental part of DSAKT is a multi-head self-attention mechanism. The "Scaled Dot-Product" proposed in \cite{Attention} describes the attention as:
\begin{equation*}
    {\rm Attention}(Q,K,V)={\rm Softmax}(\frac{QK^T}{\sqrt{d}})V
\end{equation*}
where $Q$, $K$, $V$ denotes the queries, keys and values respectively.

In the KT concept, our model should consider only first $t$ interactions when predicting the result of $(t+1)$st exercise. Therefore, we need a masking mechanism to prevent the current position from noticing the subsequent position. We use an upper triangular to mask part of the  matrix $QK^T$ with $-\infty$, which makes the attention weight of the current position to the subsequent position return to zero after softmax operation.
\begin{equation*}
    {\rm head_i= Softmax(Mask}(\frac{Q_iK_i^T}{\sqrt{d}}))V_i
\end{equation*}

In order to jointly attend to the information from different representative subspaces, we use different linear projection matrices to project the same query, key and value $h$ times.
\begin{equation*}
    {\rm Multihead}(Q,K,V)={\rm Concat(head_1,\dots,head_h)}W^O
\end{equation*}

\paragraph{Feed-Forward Networks}
The output vector obtained from the multi-head layer is still a linear combination of its inputs. Therefore, the feed-forward networks, which consists of two linear transformations with a ReLU activation in between, is used to add the non-linearity to the model.
\begin{equation*}
{\rm FFN}(x)={\rm max}(0, xW_1+b_1)W_2+b_2
\end{equation*}

\paragraph{Residual Connection \& Layer Normalization}
The residual connection proposed in \cite{resdual} are used to propagate lower level features to higher levels, which making it easier for the model to take advantage of low-level information.

The results in \cite{LN} show that the normalized information across features helps to stabilize and accelerate the neural network. For the same purpose, the layer normalization is applied in our architecture.

We apply those two structures after each sub-layer, which means that the output of each sub-layer is actually ${\rm LayerNorm}(x+{\rm Sublayer}(x))$.

\paragraph{Encoder}
The encoder consists of a multi-head attention layer with an upper triangular mask followed by a feed-forward network. Formally, the encoder can be presented as:
\begin{equation*}
\left\{
\begin{aligned}
y&={\rm LayerNorm}(x+{\rm Multihead}(x,x,x))\\
z&={\rm LayerNorm}(y+{\rm FFN}(y))\\
&{\rm Encoder}(x)=z
\end{aligned}
\right.
\end{equation*}

\paragraph{Decoder}
Similar to the encoder, the decoder also contains a multi-head attention layer and a feed-forward network.

Unlike transformer's design\cite{Attention}, our multi-head attention layer will be used twice with the same weight in the calculation. For the second time, the keys and values are the same vectors obtained from the encoder. The computation of decoder layer can be summarized as follows:
\begin{equation*}
\left\{
\begin{aligned}
y&={\rm LayerNorm}(x_1+{\rm Multihead}(x_1,x_1,x_1))\\
z&={\rm LayerNorm}(y+{\rm Multihead}(y,x_2,x_2))\\
w&={\rm LayerNorm}(z+{\rm FFN}(z))\\
&{\rm Decoder}(x_1,x_2)=w
\end{aligned}
\right.
\end{equation*}

\subsection{Prediction Layer}
Finally, each row of the vector $\tilde{r}$ obtained above passes through the linear transformation layer with sigmoid activation to obtain the predicted value of response to the exercise.
\begin{equation*}
\hat{r}={\rm Sigmoid}(\tilde{r}W+b)
\end{equation*}
\subsection{Network Training}
The purpose of the training is to minimize the negative logarithm likelihood of student response sequence observed in the model. The parameters are learned by minimizing the binary cross entropy loss between $r$ and $\hat{r}$
\begin{equation*}
\mathcal{L}=-\sum_{i}^k(r_{i}\log (\hat{r_i}) + (1-r_i)\log{(1-\hat{r_i})})
\end{equation*}

\section{EXPERIMENTS}
\subsection{Dataset}

\begin{table}[H]
\centering
\caption{Dataset Statistics}
\begin{tabular}{@{}lllll@{}}
\toprule
Dataset     & Users & Skills & Interactions & Density  
\\ \midrule
Pintia 1     &3476 &70 &142K &0.585\\
Pintia 2     &5816 &188 &225K &0.206\\
Pintia 3     &1961 &213 &108K &0.260\\
Pintia 4     &5354 &213 &316K &0.278\\
Pintia 5     &6972 &265 &591K &0.320\\
ASSIST09   &4219 &26688 &525K &0.005 \\
ASSIST12   &46676 &179999 &6123K &0.001\\
\bottomrule
\end{tabular}
\end{table}

We trained the DSAKT model based on the following datasets. On each dataset we measure area under the curve (AUC). We compare our results to Deep Knowledge Tracing.

\begin{itemize}
\item 
Pintia (1-5): These datasets are all from PTA, a platform for the Online Judge. These data sets are dense, with densities ranging from 0.2 to 0.6, as shown in Table 1. The Pintia datasets are characterized by a large number of "programming problems". Students may submit their answer repeatedly until they get it right.

\item 
ASSISTment 2009\footnote{\url{https://sites.google.com/site/assistmentsdata/home/assistment-2009-2010-data}}: The dataset is sparse as the density of this dataset is 0.005, shown in Table 1.

\item 
ASSISTment 2012\footnote{\url{https://sites.google.com/site/assistmentsdata/home/2012-13-school-data-with-affect}}: This dataset is the largest we used, but it's the most sparse, with the density of 0.001, shown in Table 1.
\end{itemize}

\subsection{Training Details}
We compare our model with the existing KT method, DKT\cite{DKT}, and SAKT\cite{SAKT}.
We implemented DSAKT with \emph{PyTorch} and the Adam optimizer\cite{Adam} with $\beta_1=0.9$, $\beta_2=0.999$ is employed to improved our model. We used the so-called Noam scheme\cite{Attention} to schedule the learning rate with $warmup\_steps=60$. Models are trained with 80\% data sets, and tested on the remaining data sets.

Batch sizes of $128$ is adopted for all datasets. We used the generally optimal hyperparameters reported in their respective papers, and set the maximum length of the sequence $k$ to be approximately proportional to the average practice per student. For Pintia, $k=50$; For ASSIST $k=100$. For DKT, we adopt hide the state dimension $d=200$ in all dataset and this of SAKT on Pintia dataset and ASSIST dataset is 24 and 64, respectively. The parameter setting of DSAKT is consistent with that of SAKT.

We trained all our models on the same machine with a NVIDIA RTX 3090 GPU.
\subsection{Results}

\begin{table}[H]
\centering
\caption{Student Performance prediction comparison.}
\begin{tabular}{@{}lllll@{}}
\toprule
            &       & \multicolumn{1}{c}{\textit{AUC}} &  \\
             \cmidrule(lr){2-4}
Dataset     & DKT   &  SAKT & DSAKT &Gain\% 
\\ \midrule
Pintia 1     &0.737 & 0.736 & 0.754	& 2.45\\
Pintia 2     &0.737 & 0.746 & 0.760 & 1.88\\
Pintia 3     &0.737 & 0.741 & 0.764 & 3.10\\
Pintia 4     &0.721 & 0.729 & 0.739 & 1.37\\
Pintia 5     &0.735 & 0.744 & 0.757 & 1.75\\
ASSIST2009   &0.746 & 0.761 & 0.769 & 1.05\\
ASSIST2012   &0.717 & 0.737 & 0.748 & 1.49\\
\bottomrule
\end{tabular}
\end{table}

The results in table 2 show that the DSAKT model performs better than SAKT and DKT model, especially on the Pintia dataset. On Pintia 3 dataset, DSAKT performs better than the competing approaches, achieving an AUC of 0.764 compared to 0.741 by SAKT, gaining a performance improvement of 3.10\%.

\begin{figure}[H]
\centering
\includegraphics[width=0.45\textwidth]{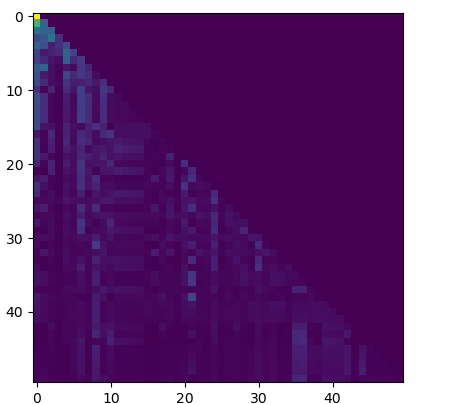}
\caption{Visualizing attention weight of encoder in a small sample of PTA data}
\end{figure}

\begin{figure}[H]
\centering
\includegraphics[width=0.45\textwidth]{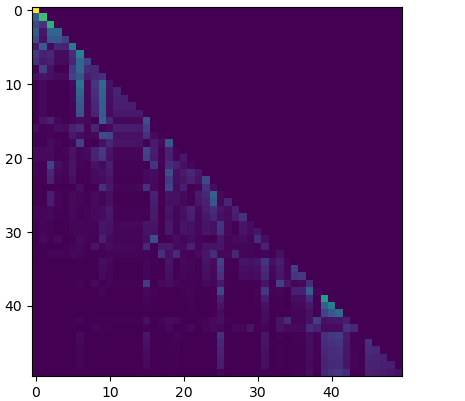}
\caption{Visualizing attention weight of decoder in a small sample of PTA data}
\end{figure}

\section{CONCLUSION} 

In this work, we proposed a new knowledge tracing model named Deep Self-Attentive Knowledge Tracing (DSAKT) and made it performed well on the data of PTA. This helps us  develop a tutoring system for all those students doing practice on PTA to provide them valuable advice. 

\bibliography{references}

\end{multicols}
\end{document}